\documentclass[10pt, a4paper]{article}
\usepackage{lrec}
\usepackage{times}
\usepackage{latexsym}
\usepackage{amsmath}
 \usepackage{hyperref}
\usepackage[T1]{fontenc}

\usepackage[utf8]{inputenc}

\usepackage{multirow}
\usepackage{graphicx}

\usepackage{gb4e}
\noautomath

\newcommand{\corpus}{{\bf FRACAS}}

\newcommand{\cue}[1]{\textcolor{teal}{[}#1\textcolor{teal}{]\textsubscript{CUE}} }
\newcommand{\speaker}[1]{\textcolor{purple}{[}#1\textcolor{purple}{]\textsubscript{SPEAKER}} }
\newcommand{\quotex}[1]{\textcolor{blue}{[}#1\textcolor{blue}{]\textsubscript{QUOTE}} }

\makeatletter
\DeclareRobustCommand\onedot{\futurelet\@let@token\@onedot}
\def\@onedot{\ifx\@let@token.\else.\null\fi\xspace}

\makeatother

\usepackage{xcolor}

\title{FRACAS: A FRench Annotated Corpus of Attribution relations in newS}

\name{\normalsize\bfseries Ange Richard\textsuperscript{1,2}\qquad Laura Alonzo-Canul\textsuperscript{1}\qquad François Portet\textsuperscript{1}\\}

\address{\small\textsuperscript{1}Univ. Grenoble Alpes, CNRS, Grenoble INP, LIG, 38000 Grenoble, France
\qquad \\ \small\textsuperscript{2}Univ. Grenoble Alpes, CNRS, Sciences Po Grenoble, PACTE, 38000 Grenoble, France  \\ 
	\footnotesize\ttfamily
	\{ange.richard, laura.alonzo-canul, francois.portet\}@univ-grenoble-alpes.fr\\
}
\abstract{Quotation extraction is a widely useful task both from a sociological and from a Natural Language Processing perspective. However, very little data is available to study this task in languages other than English. In this paper, we present a manually annotated corpus of 1676 newswire texts in French for quotation extraction and source attribution. We first describe the composition of our corpus and the choices that were made in selecting the data. We then detail the annotation guidelines and annotation process, as well as a few statistics about the final corpus and the obtained balance between quote types (direct, indirect and mixed, which are particularly challenging). We end by detailing our inter-annotator agreement between the 8 annotators who worked on manual labelling, which is substantially high for such a difficult linguistic phenomenon.}

\begin{document}

\maketitleabstract

\section{Introduction}
\label{sec:intro}

    \paragraph{}{Automatic quotation extraction and source attribution, namely finding the source speaker of a quote in a text is a widely useful, however overlooked, task: it has many applications, both on a Social Science perspective (for example, for fact-checking, detection of fake news or tracking the propagation of quotes in the news) and on a Natural Language Processing perspective (it can be tackled as a text classification task or a relation extraction task, can entail coreference resolution as well). It is however a complex task, both to define and to solve and as such has not been widely researched in NLP. There is little available corpus in English \cite{pareti-database-2012,papay-riqua-2020}, and none for French which is the language we aim to study here.}
    \paragraph{}{In this article, we contribute to the study of quotation extraction and source attribution 
        by making available \textit{FRACAS}, a human-annotated corpus of 10 965 attribution relations (quotes attributed to a speaker), annotated over a set of 1676 newswire texts in French. This corpus contains labelled information on quotations in each text, their cue and their source as well as the speaker's gender. Details on how to request the data can be found in section \ref{sec:req_data}}

\section{Related work on quotation extraction}
\label{sec:litt}
     
     \subsection{Task definition}
        \label{sec:task_def}

        \paragraph{}{Quotation is not a straightforward linguistic phenomenon, and has not been widely studied for French, although there has been a renewal of interest for its study in recent years. There is thus very little available corpora to work with, and no consensus on what the task of quotation extraction entails. The task understood as {\bf quotation extraction} aims to detect text spans that correspond to the content of a quote in a text. This task, although a seemingly straight-forward one, is deceptively simple: a quote might be announced by a cue, and may or may not be enclosed between quotation marks -- sometimes, it also contains misleading quotation marks. Its span can be very long and discontinuous, or overlap with a cue element. Quotations are usually divided into three types: direct (enclosed in quotation marks), indirect (paraphrase), and mixed or partially indirect (a combination of both). We describe these types in detail in section \ref{sec:annotation}. All these types of quotations can be found indiscriminately within different types of texts, in literary texts or in news documents, which makes their automatic identification all the more difficult.}
        
        \paragraph{}{A subsequent task to quotation extraction is the identification of the quoted speaker: this task is known as {\it source attribution}. It involves identifying for each quotation its source entity. The task is thus not only one of sequence classification ({\bf quotation extraction}) but also one of relation extraction ({\bf source attribution}) if it also involves identifying the speaker of each quote.}

        \subsection{State of the art}
    \label{sec:sota}


        \paragraph{}{The literature contains many different takes on this same task, as is described in \cite{scheible-model-2016} and \cite{vaucher-quotebank-2021}'s states of the art. There are different levels of granularity to tackling this phenomena. A few works define quotation extraction as a sentence classification task: in \cite{brunner-automatic-2013,zulaika-measuring-2022}, for example, the aim is to train a classifier to identify if a sentence input contains a quote, without emphasis on detecting the boundaries of the quote in itself.} 
        
        \paragraph{}{Other works look at the task as one of text sequence classification: they focus on extracting quote content as entities from within a text. Some of these works only consider direct quotes, which are much easier to detect due to the almost constant presence of quotations marks surrounding them. Some other works adopt a wider definition of quotations and include indirect and mixed quotes. Contributions in this line largely adopt a rule-based approach using patterns to identify quotes, speech verbs gazetteers and syntactic pattern recognition \cite{pouliquen-automatic-2007,salway-quote-2017,tu-automatic-2019,soumah-radar-2023}. It has to be noted that some works choose to apply neural architectures to detect quotations: 
        \cite{scheible-model-2016} use a pipeline system that first detects cues then quotations and combines a perceptron model and a semi-Markov model, while \cite{papay-quotation-2019} use an LSTM-based approach to detect quotation spans.}
        
        \paragraph{}{More complex approaches consider the task as a relation extraction task and seek to not only detect quote entities and cues, but to link these to their speaker entities. Up until recently, the state-of-the-art system for quotation extraction for English was the one developed by \cite{pareti-attribution-2015}, which uses a pipeline system: it first extracts cue entities with a \textit{k}-NN classifier, then uses a linear-chain conditional random field (CRF) to extract quotation spans in the close context of each cue. The results are then used as an input to a logistic regression speaker attribution model developed in \cite{okeefe-examining-2012}. It has to be noted too that a few work like that of \cite{okeefe-examining-2012,almeida-joint-2014} focus only on source attribution and co-reference resolution, as the direct speaker of a quote will sometimes be a pronoun. In these cases, a disambiguisation step has to be performed so as to determine the original source entity.}

        \paragraph{}{Our own approach is most similar to works inspired by \cite{pareti-attribution-2015}'s work, in the sense that we consider all types of quotations (direct, indirect and mixed) and seek to perform both the task of quote extraction and of source attribution. \cite{papay-quotation-2019} describe several systems with a similar goal. These systems are for the most part either rule-based or neural network-based, are trained on English data, and predate the development of BERT-like large language models which are now widely used to solve a various range of NLP tasks.}


        
        \paragraph{}{As for many complex NLP tasks, only a few system exist for for other languages (\cite{tu-etal-2021-exploration} for German, \cite{salway-quote-2017} for Norwegian, \cite{sarmento-automatic-2009} for Portuguese). For French, existing work is very scarce. Previous work on automatic quotation extraction for French date back from more than a decade ago and are mostly systems based on syntactic rules and lexicons \cite{pouliquen-automatic-2007,poulard-reperage-2008,de-la-clergerie-extracting-2009,sagot-lexicon-2010}.}

    \subsection{Available corpora}
    \label{sec:previous_corpora}
    
    \paragraph{}{Since quotation extraction is not one of the most explored NLP task, few manually labelled corpora are available for training and evaluation.} 
    
    \paragraph{}{The PARC3 English corpus \cite{pareti-database-2012} was one of the early corpora for quote extraction. It comes as an additional layer to the Penn TreeBank corpus, which is not freely available itself. 
    Since PARC3, more recent corpora have been released for English. For instance, PolNeAR v.1.0.0 \cite{newell-etal-2018-attribution} is composed of articles by 7 U.S.A. national news outlets, all covering the U.S.A. General Election campaigns of 2016. These 1008 articles are annotated with source, cue or content labels. It is also worth mentioning the SUMREN benchmark \cite{SumREN_2023} which contains 745 texts from 4 news sources that were annotated with respect to reported statements. However they contain no annotation of relations. To our knowledge, the largest existing dataset is Quotebank \cite{Quotebank} which contains 178 million articles from the Spinn3r news corpus that were automatically annotated with quotes using Quobert, a BERT based model designed by the authors to extract direct and indirect quotations as well as perform speaker attribution.
    To our knowledge, the RiQuA corpus \cite{papay-riqua-2020} is the only freely available corpus which has been deeply manually annotated  for quotation extraction and source attribution, but it only focuses on literary texts in English.
}

    \paragraph{}{In languages other than English, corpora become scarce. We can mention the RWG corpus \cite{brunner-automatic-2013} a collection of German narrative text from the 1787--1913 period annotated in direct, indirect, free indirect, and reported variants of speech. 
    \cite{zulaika-measuring-2022}'s sentence classification corpora for Spanish and Basque are also available, but do not contain information about speakers, cues and quote boundaries.  As for French, we were not able to find any freely available  labelled corpus on French quotations.
    }

    \paragraph{}{In this article we describe the \textit{FRACAS} corpus, the first freely available corpus for quotation extraction and source attribution for French. The corpus contains 10 965 attribution relations over 1676 newswire texts. The labelled entities are direct, indirect and mixed quotations. Each is attributed to their source speaker and, optionally, are linked to a cue that introduces the content of the quote. We chose newswire texts as they are more likely to contain many examples of quotations, as journalistic writing is most often based on pieces of reported speech that the journalist collected during their reporting \cite{nylund-quoting-2003}. This corpus also contains coreference annotations for quotation speakers -- namely, when the source speaker is a pronoun. We detail the contents of the corpus and the annotation process in the following section.}
       
\section{Corpus and annotations}
\label{sec:corpus}

    \subsection{Data}

    \paragraph{}{To be able to produce an annotated corpus of French news articles, our first task was to find a corpus free to use and to redistribute. There are several French news articles or newswires corpora available for research but most of them are very lightly documented as to their sources. Other better documented corpora are not free nor available to redistribute. We chose to use the Reuters Corpora 2 (RCV2), a multilingual corpus of newswires from the British news agency Reuters. These newswires were published between 1996, August 20th and 1997, August 19th, and the corpus was made available in 2005. The multilingual version contains 487 000 newswires written in 13 different language by local journalists -- it was not produced by automatic translation. This corpus is freely distributed upon request by the National Institute of Standards and Technology (NIST) \cite{reuters_nist}
    of the United States and is originally used for document classification tasks.}

    \paragraph{}{Our goal was to produce around 1500 annotated documents, each document annotated by two annotators. 1500 documents were thus randomly picked from the total 85 710 documents of the French part of the corpus. We applied on each drawn document our baseline system for quotation extraction, a rule-based algorithm by Simon Fraser University's {\it Discourse Lab} \cite{soumah-radar-2023} that we present in more details below, to make sure that the final corpus was only made of documents containing at least one quote.}

    \paragraph{}{Another batch of documents were later added to our original corpus, as after the first round of annotation, we observed that the gender ratio between quotes by men and quotes by women was highly unbalanced (only 5,4\% of speakers were Women, while Men comprised 79,9\% of annotated Speakers, the rest being labelled as Organization, Unknown or Mixed). This was a problem as our end goal for this task is to use our quote extraction model to measure gender imbalance in the news. We chose to pick another 160 files to raise the number of quotes by women. The final gender ratio is unfortunately still far from being balanced, as shown in Table~\ref{tab:info_gender}\footnote{An additional "Other" gender tag was also available for entities whose gender did not fall in the Male/Female binary. This was originally intended for cases of non-binary Agent speakers, but these were absent from our corpus, most likely due to the date of the documents. This label ended being used only 13 times out of the whole corpus by annotators, all to tag ambiguous cases of non-human Agent entities like ``un sondage'' (a poll) or ``les premiers pas de l'enquête'' (the detective's first findings). We chose to remove them from this table for space and relevancy reasons.}. The low presence of quoted women in the newswires might be explain by the fact that in the mid-1990s women were even more scarce in media than today. The final corpus contains 1676 documents, divided into train, dev and test sets as shown in Table~\ref{tab:info_corpus}.}

    \begin{table}
    \centering
      \begin{tabular}{|c|c|c|c|}
      \hline
        \bf{Partition} & \bf{\# docs} & \bf{\# tokens} & \begin{tabular}[c]{@{}l@{}}\bf{Mean \# tokens} \\ {\bf per doc}\end{tabular} \\
        \hline
        train & 1114 & 436 150 & 391.51 \\
        dev & 281 & 131 202 & 466.91 \\
        test & 281 & 137 083 & 487.83 \\ \hline
        TOTAL & 1676 & 704 435 & 448.75 \\
        \hline
      \end{tabular}
        \caption{Number of documents and tokens in the \corpus\ corpus \label{tab:info_corpus}}
    \end{table}

\begin{table*}[]
\begin{small}
\begin{tabular}{|llll|llll|l|}
\hline
\multicolumn{1}{|l|}{\multirow{2}{*}{}} & \multicolumn{3}{|l|}{\textbf{QUOTES}} & \multicolumn{4}{|l|}{\textbf{\begin{tabular}[c]{@{}l@{}}SPEAKER\\ (as direct source / as coreferent)\end{tabular}}}& \multirow{2}{*}{\textbf{CUE}} \\
\multicolumn{1}{|l|}{} & \multicolumn{1}{l|}{Direct} & \multicolumn{1}{l|}{Indirect} & Mixed & \multicolumn{1}{l|}{Agent} & \multicolumn{1}{l|}{Organization} & \multicolumn{1}{l|}{GoP} & \multicolumn{1}{|l|}{SP} & \\ \hline
\multicolumn{1}{|l|}{train} & \multicolumn{1}{l|}{2698(40)} & \multicolumn{1}{|l|}{2919(43)} & 1123(17) & \multicolumn{1}{l|}{1881(34) / 1196(91)} & \multicolumn{1}{l|}{1071(20) / 83(6)} & \multicolumn{1}{l|}{628(11) / 38(3)} & 1928(35) & 6439 \\
\multicolumn{1}{|l|}{dev} & \multicolumn{1}{l|}{820(40)} & \multicolumn{1}{l|}{835(41)} & 404(19) & \multicolumn{1}{l|}{570(40) / 312(84)} & \multicolumn{1}{l|}{386(27) / 43(11)} & \multicolumn{1}{l|}{158(11) / 17(5)} & 312(22) & 1978 \\
\multicolumn{1}{|l|}{test} & \multicolumn{1}{l|}{919(42)} & \multicolumn{1}{|l|}{918(41)} & 382(17) & \multicolumn{1}{l|}{628(41) / 353(86)} & \multicolumn{1}{l|}{360(24) / 36(9)} & \multicolumn{1}{|l|}{191(12) / 21(5)} & 353(23) & 2126 \\ \hline
\textbf{TOTAL} & \multicolumn{1}{|l|}{4437(40)} & \multicolumn{1}{l|}{4672(43)} & 1909(17) & \multicolumn{1}{l|}{3079(41) / 1861(89)} & \multicolumn{1}{|l|}{1817(25) / 162(8)} & \multicolumn{1}{l|}{677(9) / 76(3)} & 1861(25) & 10543 \\
\hline
\end{tabular}
\end{small}

\caption{Number (and percentage per partition) of annotations per entity and relation label in \textit{FRACAS} (GoP = Group of People, SP = Source Pronoun)}
\label{tab:annotation_info}
\end{table*}

\begin{table}
    \centering
      \begin{tabular}{|c|c|c|c|c|}
        \hline
        \bf{Partition} & \bf{Male} & \bf{Female} & \bf{Mixed} & \bf{Unknown} \\
        \hline
        train & 2706(79) & 257(7) & 132(4) & 336(10) \\
        dev & 613(58) & 246(23) & 92(9) & 101(10) \\
        test & 705(60) & 254(21) & 110(9) & 117(10) \\ \hline
        TOTAL & 4024(71) & 757(13) & 334(6) & 554(10) \\
        \hline
      \end{tabular}      
    \caption{Gender distribution of speaker entities (Agents and Group of People only) \label{tab:info_gender}}
\end{table} 
    
    \subsection{Annotation process}
    \label{sec:annotation}

    \paragraph{Annotation guidelines.}{We draw on \cite{pareti-database-2012}'s work for the annotation guidelines and labels. We consider a quote as a triplet made of three entities: a quote content, a speaker and a cue (this last entity is optional but most quotes are introduced by a cue, very often a verb). We distinguish quote types and speaker types. Quote types are the following:}
    
    \begin{itemize}
        \item {\bf Direct Quotation}: a direct quotation reports the quoted speaker's exact words. It is the easiest type to spot as it is usually enclosed by quotation marks.
        \item[]\begin{exe}
            \ex \speaker{Nicki}  \cue{said} \quotex{``Let's go to the beach!''}.
        \end{exe}
         \item {\bf Indirect Quotation}: an indirect quotation is a paraphrase of the speaker's words. It is usually a rephrasing of these words, and is most often written in the 3rd person, without quotation marks.
         \item[]\begin{exe}
            \ex \speaker{Rihanna}  \cue{asked} \quotex{not to stop the music}.
        \end{exe}
        \item {\bf Mixed Quotation}: a mixed quotation is a paraphrase (indirect quotation) that contains direct speech elements (words or part of a sentence), usually enclosed within quotation marks.
        \item[]\begin{exe}
            \ex \speaker{Britney} \cue{said} that \quotex{she did it ``again.''}
        \end{exe}
    \end{itemize}
    
    \paragraph{}{We divide Speaker labels as following: {\bf Agent} (when the speaker is a single person, i.e ``Mariah Carey''), {\bf Group of People} (i.e ``The cast of {\it Drag Race France}''), {\bf Organization} (i.e ``the UNESCO''), or {\bf Source Pronoun} (i.e ``she''). In that last case, the pronoun is linked to its referent entity, labelled with one of the Speaker tags, as in the following example\footnote{Note that in this example, the first sentence is not considered as a quote, even with the presence of the ambiguous speech verb ``warned''}.}
    
    \begin{exe}
        \ex \textcolor{purple}{[}Beyoncé\textcolor{purple}{]\textsubscript{SPEAKER (Agent)}} warned him! \textcolor{purple}{[}She\textcolor{purple}{]\textsubscript{SPEAKER (Source pronoun)}} \cue{told} him \quotex{he should have put a ring on it}. 
    \end{exe}
    
    \paragraph{}{Additionally, each speaker is labelled with a gender tag amongst the following: Male, Female, Mixed, Other or Unknown. The gender was assigned based on linguistic features like gender agreement and other semantic references found within the text. Each quote is linked by a relation to a speaker and a cue, and each source pronoun is linked to a referent labelled with one of the above speaker labels. The overall numbers of tagged entities for each label is detailed in table \ref{tab:annotation_info}.}

\begin{figure*}[htb]
    \centering
        \includegraphics[width=\linewidth]{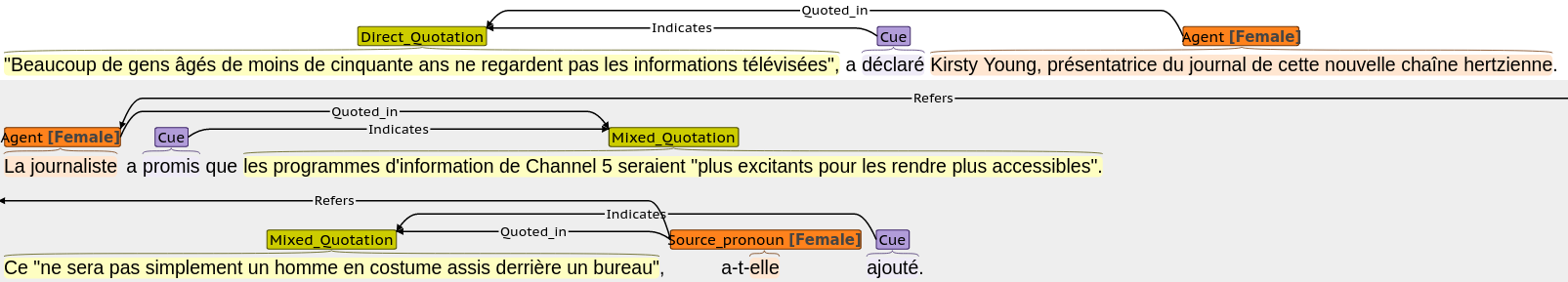}
        \caption{Screenshot of an annotated text in the BRAT interface}
        \label{fig:brat_screenshot}
    \end{figure*}

    \paragraph{Annotators.}{We used the software BRAT \cite{brat-annotation-tool} for this annotation task, as shown in Figure \ref{fig:brat_screenshot}. The original corpus (1436 newswires) were annotated by a team of 9 annotators: 7 women, 1 men and 1 non-binary person, all graduate students from NLP, Communication Studies or Linguistic degrees recruited through an open call for temporary workers through the university channels. The annotators were paid according to the French minimum wage (a gross salary of 18,75€ per hour) for a 20-hour long contract and each had to annotate 300 documents. The documents were split by batches of 150 and each batch was annotated by 2 different annotators. The annotators received detailed annotation guidelines and half a day of online training with the annotation campaign supervisor to make sure that the guidelines and the use of the software were understood. The annotators had about a month to complete the annotation task, between June and July 2021. At halfway point, the ongoing annotation were checked by the supervisor and individual feedback was sent to annotators to clarify misunderstood instructions.}

    \paragraph{}{The additional 168 documents that were later added to raise the number of quotes by women were annotated by three experts annotators: two researchers from the project and one of the annotators previously trained for the campaign. For this subsequent annotation campaign, the documents were divided amongst the three annotators, with an intersection of 30\% of the documents that were annotated by all three annotators to calculate inter-annotator agreement.}
            
        \subsubsection{Inter-Annotator Agreement (IAA)}


        \paragraph{}{After both annotation campaigns, the annotated documents were cleaned and preprocessed before computing IAA to correct some easy-to-fix annotation errors. For instance, we automatically edited annotated spans that included wrong boundaries such as extra punctuation or white spaces, or were missing elements such as direct quotes entities that did not include their quotation marks. We also noticed that some guidelines hadn't been understood by all annotators: the instructions were to annotate any elements that were syntactically linked to a speaker (i.e: in a speaker phrase like ''Madonna, queen of pop``, the whole phrase should be annotated as a Speaker and not only ''Madonna``). However some annotators did not annotate all the elements. We chose to reprocess these instances by keeping the longest version of an annotated phrase if two Speaker entities were overlapping between two annotators.}
        
        \paragraph{}{To measure Inter-annotator agreement for entity annotation, we use the $\gamma$ score proposed by \cite{mathet-unified-2015}. Since our annotation task was one of sequence delimitation and classification, the $\gamma$ score allows to compute an agreement that accounts for both unitizing (agreement on unit span location within a text) and categorization (agreement on unit labeling). The $\gamma$ score, like Cohen's $\kappa$, is computed from observed and expected disagreements, instead of agreements. We consider it the best IAA score for our task as it takes into account overlap when calculating unit alignment, chance correction and category weight (disagreements on rare categories are more serious than on frequent ones). The results for all entities is showcased on Table \ref{tab:iaa}. We obtain an inter-annotator agreement score of 0.76 on all entities, which is satisfying considering the difficulty of the task at hand. We observe the best IAA scores for Direct Quotation and Source pronoun entities, as well as for Cues and Agent speakers. Indirect Quotation annotations obtain the lowest score. We link this to the difficulty to determine what is a paraphrase or not, and what is to be included in the span of the quotation. These score allow us to note that the quotation detection task, when extended to its larger comprehension (meaning including indirect and mixed quotations), is not an easy one even by human standards.}
        
        \paragraph{}{To determine which annotations should be included in the final corpus, we computed an IAA with relation to a gold standard for each batch of 150 document annotated by a pair of annotator. The gold standard was composed of 10 documents of each batch annotated by an expert annotator. An IAA score was computed with the same $\gamma$ measure for each annotator for each batch over the 10 documents. The documents annotated by the annotator who had the highest IAA with the gold standard were then kept in the final corpus.}

\begin{table}
            \centering
              \begin{tabular}{c|c}
                \bf{Entity label} & \bf{$\gamma$ agreement} \\
                \hline
                \bf{All entities} & 0.7699\\
                 Direct Quotation (Q) & 0.8857 \\
                Indirect Quotation (Q) & 0.6415 \\
                 Mixed Quotation (Q) & 0.7468 \\
                Cue & 0.8291 \\
                Agent (S) & 0.8337 \\
                Organization (S) & 0.7858 \\
                Group of people (S) & 0.7828 \\
                Source pronoun (S) & 0.898 \\
              \end{tabular}
              \caption{$\gamma$ agreement between annotators of first annotation campaign (S for Speaker entity types, Q for Quotation entity types)}
            \label{tab:iaa}
            \end{table}

\section{Conclusion}
\label{sec:ccl}
    \paragraph{}{In this article we detailed the annotation process behind the FRACAS corpus, a corpus for Quotation Extraction and Source attribution for French. We chose to build our annotated corpus from the newswire texts from the RCV2 Reuters corpus distributed by the NIST in order to make this corpus freely available. As such, both the original text (from NIST) and the annotations (from us) is available. We adopted an extensive definition of quotations, as we include Direct, Indirect and Mixed Quotes, and obtained a total of 10 965 annotated attribution relations spanning over 1676 texts. Our annotation process involved 8 annotators and yields satisfying inter-annotator agreement results, which also underlines the complexity of this phenomenon. We make this corpus available, and will use our data in following work to train quotation extraction and attribution systems for French with state-of-the-art architectures. We plan to use these systems to measure gender imbalance in quotations in French media. Quotation extraction is also essential to track who said what in the press and can be useful for the detection of misquotations and fake news.}

\section{Request of data}
\label{sec:req_data}
    Please visit our Zenodo repository to find the instructions to request the data: \url{https://zenodo.org/record/8353229}

\section{Acknowledgements}
This work is part of a project funded by an Initiative de recherche à Grenoble (IRGA ANR-15-IDEX-02) project from the Université Grenoble Alpes, and has been partily supported by MIAI@Grenoble-Alpes (ANR-19-P3IA-0003).

\section*{Bibliographical References}
\label{main:ref}
\bibliographystyle{lrec}
\bibliography{references}

\end{document}